\documentclass[11pt,letterpaper]{article}

\usepackage[
  letterpaper,
  top=1in,
  bottom=1in,
  left=1in,
  right=1in
]{geometry}

\usepackage{libertinus}
\usepackage{libertinust1math}
\usepackage{amsthm}
\PassOptionsToPackage{hyphens}{url}

\makeatletter

\@ifpackageloaded{inputenc}{}{\usepackage[utf8]{inputenc}}
\@ifpackageloaded{fontenc}{}{\usepackage[T1]{fontenc}}
\@ifpackageloaded{natbib}{}{\usepackage[numbers]{natbib}}
\@ifpackageloaded{url}{}{\usepackage[hyphens]{url}}
\@ifpackageloaded{hyperref}{}{\usepackage{hyperref}}

\@ifpackageloaded{booktabs}{}{\usepackage{booktabs}}
\@ifpackageloaded{threeparttable}{}{\usepackage{threeparttable}}
\@ifpackageloaded{makecell}{}{\usepackage{makecell}}
\@ifpackageloaded{array}{}{\usepackage{array}}
\@ifpackageloaded{rotating}{}{\usepackage{rotating}}
\@ifpackageloaded{multirow}{}{\usepackage{multirow}}
\@ifpackageloaded{multicol}{}{\usepackage{multicol}}

\@ifpackageloaded{amsfonts}{}{\usepackage{amsfonts}}
\@ifpackageloaded{amsmath}{}{\usepackage{amsmath}}
\@ifpackageloaded{amssymb}{}{\usepackage{amssymb}}
\@ifpackageloaded{mathtools}{}{\usepackage{mathtools}}
\@ifpackageloaded{amsthm}{}{\usepackage{amsthm}}
\@ifpackageloaded{nicefrac}{}{\usepackage{nicefrac}}

\@ifpackageloaded{graphicx}{}{\usepackage{graphicx}}
\@ifpackageloaded{subfigure}{}{\usepackage{subfigure}}
\@ifpackageloaded{wrapfig}{}{\usepackage{wrapfig}}
\@ifpackageloaded{caption}{}{\usepackage{caption}}

\@ifpackageloaded{algorithm}{}{\usepackage{algorithm}}
\@ifpackageloaded{algorithmicx}{}{\usepackage{algorithmicx}}
\@ifpackageloaded{algpseudocode}{}{\usepackage{algpseudocode}}

\@ifpackageloaded{microtype}{}{\usepackage{microtype}}
\@ifpackageloaded{xcolor}{}{\usepackage[table]{xcolor}}
\@ifpackageloaded{tcolorbox}{}{\usepackage[most]{tcolorbox}}
\@ifpackageloaded{adjustbox}{}{\usepackage{adjustbox}}
\@ifpackageloaded{enumitem}{}{\usepackage[inline]{enumitem}}
\@ifpackageloaded{placeins}{}{\usepackage{placeins}}
\@ifpackageloaded{inconsolata}{}{\usepackage{inconsolata}}
\@ifpackageloaded{cleveref}{}{\usepackage[capitalize,noabbrev]{cleveref}}

\@ifpackageloaded{pgfplots}{}{\usepackage{pgfplots}\pgfplotsset{compat=1.17}}

\makeatother

\hypersetup{
  colorlinks=true,
  linkcolor={blue!70!black},
  citecolor={blue!70!black},
  urlcolor={blue!70!black},
  pdftitle={Self-Improving In-Context Learning},
  pdfauthor={Baturay Saglam, Dionysis Kalogerias},
}

\DeclareMathOperator*{\argmax}{arg\,max}

\definecolor{customblue}{HTML}{1F77B4}
\definecolor{customdarkblue}{HTML}{0000FF}
\definecolor{customdarkred}{HTML}{FF0000}
\definecolor{customorange}{HTML}{FF7F0E}
\definecolor{customgreen}{HTML}{2CA02C}
\definecolor{customred}{HTML}{D62728}
\definecolor{custombrown}{HTML}{8C564B}
\definecolor{custompurple}{HTML}{9467BD}
\definecolor{indigo}{HTML}{000080}
\definecolor{darkgreen}{HTML}{3C4C24}
\definecolor{darkorange}{HTML}{D15300}
\definecolor{coral}{HTML}{E64A19}
\definecolor{teal}{HTML}{00796B}

\definecolor{customorange}{HTML}{FF7F0E}

\newtcbox{\highlight}[1][customorange]{
    on line,
    arc=0pt,
    colback=#1!35!white,
    colframe=#1!35!white,
    boxsep=0pt,
    left=1pt,
    right=1pt,
    top=2pt,
    bottom=2pt,
    boxrule=0pt,
    nobeforeafter
}

\newcommand{\best}[1]{{\color{black!80}\textbf{#1}}}

\usepackage{pifont}
\newcommand{\cmark}{\ding{51}}
\newcommand{\xmark}{\ding{55}}



\theoremstyle{plain}

\theoremstyle{definition}

\theoremstyle{remark}

\title{%
  \vspace{-0.5in}%
  \textbf{\Large Self-Improving In-Context Learning}%
  \vspace{0.08in}%
}

\author{%
  {\large Baturay Saglam \quad Dionysis Kalogerias}\\[1.2em]
  {\normalsize Department of Electrical and Computer Engineering}\\
  {\normalsize Yale University}\\[0.35em]
  {\small \texttt{\{\href{mailto:baturay.saglam@yale.edu}{baturay.saglam}, \href{mailto:dionysis.kalogerias@yale.edu}{dionysis.kalogerias}\}@yale.edu}}%
}
\date{}

\begin{document}

\maketitle

\begin{abstract}
\noindent
We propose to improve in-context learning (ICL) by optimizing the continuous embeddings of a fixed few-shot prompt at test time. The key observation is that the log-probabilities a model assigns to its demonstrated outputs---available from a single forward pass without generating any tokens---provide a meaningful signal for how well the model has inferred the task from its demonstrations. We formalize this signal as a bounded, self-supervised confidence proxy and maximize it via zeroth-order optimization over the prompt embeddings, yielding a test-time calibration procedure. The approach requires no finetuning, no token generation, no predefined label set, and no external data, making it equally applicable to both classification and free-form generation tasks. Across a comprehensive suite of ICL tasks, the proposed calibration consistently matches or improves upon the base model and outperforms classification-specific baselines on most tasks. The statistically significant correlation between proxy improvement and downstream accuracy gain confirms that the proposed proxy encodes a reliable optimization signal for in-context learning.
\end{abstract}

\vspace{0.1in}

%%%%%%%%%%%%%%%%%%%%%%%%%%%%%%%%%%%%%%%%%%%%%%%%%%%%%%%%%%%%%%%%%%%%%%%%%%%%%%%%
% MAIN CONTENT
%%%%%%%%%%%%%%%%%%%%%%%%%%%%%%%%%%%%%%%%%%%%%%%%%%%%%%%%%%%%%%%%%%%%%%%%%%%%%%%%
\section{Introduction}
\label{sec:introduction}

In-context learning (ICL) enables large language models (LLMs) to perform new tasks by conditioning on a small set of input--output demonstrations in the prompt, without updating model parameters~\citep{gpt3}. This capability has made few-shot prompting a dominant paradigm for deploying language models~\citep{min_2022}. However, it is also notably fragile: even with semantically identical demonstrations, simply reordering them can shift accuracy from near state-of-the-art to near random~\citep{lu_ordering}. Performance is similarly sensitive to the choice and formatting of examples. This brittleness has motivated a growing body of work on improving the reliability of ICL at inference time.

Existing test-time methods fall into three broad categories: demonstration selection, demonstration ordering, and output calibration. Each family has shown practical improvements in its target setting, yet all share a common structural limitation: they operate either on discrete prompt-level decisions (i.e., which examples to include and in what order) or on the model's output probabilities after the fact. Most additionally require a finite, predefined label set, restricting them to classification (see Appendix~\ref{sec:excluded_baselines}). We explore a complementary direction.

A fundamentally different strategy is to intervene on the continuous prompt embedding matrix that the model directly conditions on. Since the model operates over this matrix rather than the discrete tokens, adjustments in embedding space can reshape the output distribution while leaving the human-readable input intact~\citep{detox, tsa}. We propose to improve ICL by optimizing the continuous embeddings of a fixed few-shot prompt. Our key observation is that the model's own log-probabilities on the demonstrated output labels---obtainable from a single teacher-forced forward pass, without generating any tokens---provide a meaningful signal for how well the model has inferred the task from its demonstrations. We formalize this signal as a scalar, bounded confidence proxy with three complementary facets: absolute predictive confidence on each demonstrated label, robustness to low-probability tokens in the output spans, and progressive improvement in prediction quality across demonstrations.

To maximize this proxy, we estimate its gradient with respect to the input embeddings via zeroth-order optimization and iteratively update the embeddings along the estimated gradient direction. The resulting procedure calibrates the prompt representation at test time, steering it toward regions of the embedding space associated with higher in-context confidence. It requires no external data, no access to model parameters, and no additionally learned parameters---only the input embeddings and the model's log-probabilities are assumed to be available. Each optimization step consists entirely of forward passes; no tokens are generated at any point, and the original discrete prompt is never modified---only its continuous embeddings are. Because the proxy is computed solely from the model's predictions on the demonstrations already present in the prompt, no predefined label set is required, making the method equally applicable to classification and free-form generation. It can therefore be freely composed with any existing demonstration selection, ordering, or calibration strategy, and applied as a plug-and-play module to any off-the-shelf autoregressive language model.

Across a comprehensive suite of classification and free-form generation tasks designed to probe rule learning and exact copying~\citep{icleval}, and over several model scales, the proposed method consistently matches or improves upon the base model while outperforming classification-specific baselines on most tasks out of the box. Furthermore, the correlation between proxy improvement and downstream accuracy gain is statistically significant across all models combined, confirming that the proxy encodes a reliable optimization signal for in-context learning. We open-source our code at \url{https://github.com/baturaysaglam/self-improving-ICL}.
\section{Related Work}
\label{sec:related_work}

\subsection{In-Context Learning}

In-context learning is sensitive to the choice of demonstrations, their ordering, and the decoding procedure. A complementary line of work studies \emph{why} ICL works---through the role of demonstration labels~\citep{min_2022,yoo_2022}, information flow through label tokens~\citep{wang_2023a}, the formation of task representations~\citep{hendel_icl, icv, todd_icl, baturay_icl}, and connections to implicit gradient descent~\citep{dai_2023, transformers_learn_icl_by_grad_des}---but lies outside our scope.

\paragraph{Demonstration selection.}
The choice of in-context examples can dramatically shift ICL performance~\citep{kate}. Existing methods retrieve nearest neighbors in a pretrained embedding space~\citep{kate,dpp_icl,vote_k}, score candidates via the target model's own feedback~\citep{sa_icl,cone,se2,lens_icl,misconfidence,ids_icl,dva_icl}, apply task-specific heuristics such as reasoning complexity or structural coverage~\citep{complexity_prompting,diverse_demos}, or have the model generate its own demonstrations~\citep{sg_icl,z_icl,usp,cosp}. All assume access to a scorable candidate pool or a finite label space.

\paragraph{Demonstration ordering.}
The order of demonstrations alone can shift accuracy by tens of percentage points~\citep{lu_ordering}. Subsequent work addresses this through instance-adaptive reordering~\citep{demo,xu_label_dist_ordering,oeoicl,cluster_search} or by eliminating order sensitivity entirely~\citep{batch_icl}; however, the scoring functions used generally require a finite label set, and permutation search scales combinatorially with the number of demonstrations.

\paragraph{Output calibration.}
A separate family corrects systematic biases---majority-label, recency, and surface-form effects---by adjusting the model's output distribution~\citep{cc,dc_pmi,noisy_channel,dc_cal,batch_cal,answer-level-calibration,task_cal,knn_prompting} or operating on internal representations~\citep{proca,hidden_cal}. All require a known label set, restricting them to classification; several further need transductive access to a batch of test inputs~\citep{batch_cal,proca,pride} or to hidden states~\citep{hidden_cal}.

Our method is orthogonal to all three families. It optimizes the continuous representation of a \emph{fixed} prompt and requires no candidate pool, test batch, label set, or access to model internals beyond the embedding layer. Its cost does not scale with the number of demonstrations, it operates identically on classification and open-ended generation, and it can be composed freely with any selection, ordering, or calibration strategy.

Appendix~\ref{sec:excluded_baselines} formalizes this comparison and explains why the listed methods are structurally incompatible with open-ended generation.

\subsection{Zeroth-Order Optimization in LLMs}
Finite-difference estimators~\citep{nesterov_zoo} have been mainly used for privacy- and memory-efficient finetuning~\citep{dpzero,sperse_mezo,mezo,mezo_svrg} and as a substitute for gradient estimation in soft prompt optimization~\citep{zot}. All of these works operate at training time under a dataset-level loss. In contrast, we operate at the instance level: the gradient is estimated for a single prompt at test time, within a small forward-pass budget, in the fundamentally different regime of in-context learning. More recently, \citet{detox, tsa} have shown that optimizing input embeddings can steer a model toward behavior satisfying certain properties. Our setting differs, however: we devise a self-supervised objective and operate in ICL, whereas they target safety properties (e.g., toxicity) for which a noise-free oracle is available from an external provider (e.g., an API). Moreover, we use input-embedding optimization as a tool to demonstrate the effectiveness of the proposed proxy, rather than as the objective of our study.
\section{Background}
\label{sec:background}

%% ─────────────────────────────────────────────
%% Autoregressive Text Generation
%% ─────────────────────────────────────────────
\subsection{Autoregressive Text Generation}
\label{sec:autoregressive}

Large language models generate text autoregressively, producing one token at a time with each prediction conditioned on all preceding tokens.
Let $\mathcal{V}$ denote a finite vocabulary of $V = |\mathcal{V}|$ tokens.
We write a text sequence of length $L$ as $x_{1:L} = (x_1, x_2, \ldots, x_L)$, where each $x_t \in \mathcal{V}$.
An autoregressive model parameterized by $\theta$ defines the joint probability of the sequence via the chain rule of probability:
\[
    P_\theta(x_{1:L}) \;=\; \prod_{t=1}^{L} P_\theta\!\left(x_t \mid x_{<t}\right),
\]
where $x_{<t} = (x_1, \ldots, x_{t-1})$ is the prefix preceding position $t$.

Each token $x_t$ is first mapped to a dense vector $e_t \in \mathbb{R}^d$ through a learned embedding matrix $E \in \mathbb{R}^{V \times d}$. The transformer architecture~\citep{transformer} processes the resulting embedding sequence and produces, at each position, a conditional distribution over the next token. The log-probability of the observed token $x_t$ is denoted
\begin{equation*}
    \ell_t(X)\; = \;\log P_\theta(x_t\mid x_{<t}),\qquad t=1,\ldots,L,
\end{equation*}
which satisfies $\ell_t \leq 0$, with equality only when the model assigns probability one to $x_t$.

When a complete sequence is provided as input---as in the prompt-based setting we consider---a single forward pass yields the full sequence of log-probabilities $(\ell_1, \ldots, \ell_L)$ simultaneously, which is central to the practicality of our approach.

%% ─────────────────────────────────────────────
%% In-Context Learning
%% ─────────────────────────────────────────────
\subsection{In-Context Learning}
\label{sec:icl}

A \emph{few-shot prompt} $\mathcal{P}$ is constructed by concatenating $T$ demonstration pairs followed by a query input:
\begin{equation}
\label{eq:icl_prompt}
    \mathcal{P} \;=\; \bigl[\,s_1,\; y_1,\; s_2,\; y_2,\; \ldots,\; s_T,\; y_T,\; s_{\mathrm{query}}\,\bigr],
\end{equation}
where each $(s_i, y_i)$ consists of a task input $s_i$ and its corresponding output $y_i$.

When tokenized, the prompt $\mathcal{P}$ becomes a token sequence $x_{1:L} = (x_1, \ldots, x_L)$. For each demonstration $i$, we denote by $\mathcal{Y}_i \subseteq \{1, \ldots, L\}$ the set of token positions corresponding to the output $y_i$. These output-span log-probabilities reflect how confidently the model predicts each demonstrated output given its context---the optimization signal of Section~\ref{sec:methodology}.

%% ─────────────────────────────────────────────
%% Zeroth-Order Optimization
%% ─────────────────────────────────────────────
\subsection{Zeroth-Order Optimization}
\label{sec:zo}

Zeroth-order methods estimate gradient information from function evaluations alone---useful when the objective is black-box or non-differentiable; we rely on the Gaussian smoothing framework of~\citet{nesterov_zoo}.

Given an objective $f\colon \mathbb{R}^n \to \mathbb{R}$ and a smoothing parameter $\mu > 0$, the \emph{Gaussian-smoothed} counterpart of $f$ is defined as
\begin{equation*}
    f_\mu(X) \;=\; \mathbb{E}_{U \sim \mathcal{N}(0, I_n)}\!\bigl[f(X + \mu U)\bigr].
\end{equation*}
If $f$ is Lipschitz-continuous, i.e., $|f(X) - f(Y)| \le L_0 \|X - Y\|$ for all $X, Y$, then $f_\mu$ is differentiable for every $\mu > 0$ and approximates $f$ with a controlled error of order $\mathcal{O}(\mu\sqrt{n})$.
We note that these conditions are invoked only to motivate the bound; in a fully black-box setting they cannot be verified from query access to $f$. Nonetheless, the resulting estimator can be applied empirically regardless of whether the underlying constants are known.

A central result of \citet{nesterov_zoo} establishes that the gradient of $f_\mu$ admits the familiar finite-difference form:
\begin{equation}
\label{eq:grad_smoothed}
    \nabla f_\mu(X) \;=\; \mathbb{E}_{U}\!\left[\frac{f(X + \mu U) - f(X)}{\mu}\,U\right].
\end{equation}
This identity requires only Lipschitz continuity of $f$; differentiability of $f$ itself is not needed. The baseline term $f(X)/\mu$ does not bias the estimate (since $\mathbb{E}[U]=0$) but reduces its variance.

In practice, the expectation in \eqref{eq:grad_smoothed} is replaced by a Monte Carlo average over $N$ independent perturbations $U_1, \ldots, U_N \sim \mathcal{N}(0, I_n)$:
\begin{equation}
\label{eq:zoo_estimator}
    \hat{g}(X) \;=\; \frac{1}{N}\sum_{i=1}^{N} \frac{f(X + \mu U_i) - f(X)}{\mu}\,U_i \;\approx\; \nabla f_\mu(X).
\end{equation}
The smoothing parameter $\mu$ governs a bias--variance tradeoff: smaller values produce a sharper approximation to $\nabla f$ but amplify the variance of the finite-difference terms, while larger values yield smoother but more biased estimates.
The number of samples $N$ controls the variance of the Monte Carlo average.

\section{Self-Improving In-Context Learning}
\label{sec:methodology}

%% ─────────────────────────────────────────────
%% Overview
%% ─────────────────────────────────────────────
\paragraph{Overview.}
\label{sec:methodology_overview}

Given a few-shot prompt as in \eqref{eq:icl_prompt}, we treat the language model as a black box that, for any provided prompt, returns teacher-forced token log-probabilities. Our key assumption is that the token positions corresponding to each demonstrated output span are known (denoted $\mathcal{Y}_1,\ldots,\mathcal{Y}_T$ in Section~\ref{sec:icl}).

Let $X \in \mathbb{R}^{L \times d}$ denote the embedding matrix of the prompt tokens, and let $f(X)$ be a scalar proxy quantifying the model's confidence on the demonstrations given the prompt context.
%
% using \emph{only} the log-probabilities of the label tokens in $\mathcal{Y}_1,\ldots,\mathcal{Y}_T$ produced by a single forward pass. 
%
% Intuitively, $f(X)$ should be large when the model assigns high probability to the demonstrated labels, when this confidence does not hinge on a small number of low-probability ``failure'' tokens, and when confidence tends to improve as more demonstrations are observed.
% We define $f$ concretely in the next subsection.
% 
We seek an embedding-space optimization that increases the proxy value:
\begin{equation*}
X^\star \;\in\; \argmax_{X}\; f(X),
\end{equation*}
with the understanding that the discrete prompt text is held fixed and only its continuous embeddings are modified; each embedding remains within its token subspace. The proxy $f$ may be non-smooth due to robust aggregation over tokens, and backpropagating through the model to obtain $\nabla_X f(X)$ is prohibitively expensive at test time. Instead, we estimate this gradient via zeroth-order optimization reviewed in Section~\ref{sec:zo} and defined in \eqref{eq:zoo_estimator}.

We form a stochastic ascent direction $\hat{g}_k~\approx~\nabla_X f_\mu(X_k)$ from evaluations of $f$ at randomly perturbed embeddings $X_k + \mu U_i$. Because each row of $U_i$ is drawn independently, each token embedding is perturbed separately, allowing the estimator to capture token-specific sensitivity of $f$. Since the proxy is to be computed from log-probabilities at demonstration positions, which under causal attention are unaffected by the query tokens that follow them, the perturbation is restricted to the demonstration region---noise at query positions is set to zero.

The embeddings are then updated iteratively as $X_{k+1} = X_k + \eta\,\hat{g}_k$, where $\eta$ is the step size. This update steers the prompt representation toward regions of the embedding space associated with higher in-context confidence. The procedure operates entirely at test time, requires only additional forward passes per input instance, and assumes access only to the input embeddings and model log-probabilities.

%% ─────────────────────────────────────────────
%% A Proxy for In-Context Learning Confidence
%% ─────────────────────────────────────────────
\subsection{A Proxy for In-Context Learning Confidence}
\label{sec:proxy}

We define the proxy $f(\cdot)$ subject to three design principles balancing theoretical constraints and practical considerations:
\begin{enumerate}[label=\emph{(D\arabic*)},leftmargin=*,itemsep=2pt,topsep=4pt]
    \item \emph{Bounded and continuous.} $f(X)\in[0,1]$, ensuring compatibility with the zeroth-order estimator and controlled approximation error.
    \item \emph{Single forward pass.} Each evaluation depends only on teacher-forced log-probabilities $\{\ell_t : t\in\bigcup_i\mathcal{Y}_i\}$ from the demonstrations already present in the prompt; the full vocabulary distribution is neither extracted nor stored, no tokens are generated, and no ground-truth label candidates for the query are required (in contrast to, e.g.,~\citealp{xu_label_dist_ordering, lu_ordering, demo}).
    \item \emph{Behaviorally informative.} $f$ increases as the model's predictions on the demonstrated outputs improve, reflecting genuine task understanding and thereby translating into downstream task performance.
\end{enumerate}
%
% Several ICL scoring functions from prior work (e.g.,~\citep{sa_icl, cone, cc, dc_pmi, oeoicl, noisy_channel}) violate one or more of these principles, as they typically require leave-one-out evaluation, token generation, or reference inputs.

%% ─────────────────────────────────────────────
% \paragraph{Setup.}
% Consider a few-shot prompt with embedding matrix $X\in\mathbb{R}^{L\times d}$ and demonstration output spans $\mathcal{Y}_1,\ldots,\mathcal{Y}_T$ as defined in Section~\ref{sec:icl}. Because we optimize over $X$, the token log-probabilities $\ell_t$ introduced in Section~\ref{sec:autoregressive} now depend on the embeddings; we make this dependence explicit by writing
% \[
% \ell_t(X)\;\coloneqq\;\log P_\theta(x_t\mid x_{<t}),\qquad t=1,\ldots,L,
% \]
% and $p_t(X)\coloneqq \exp(\ell_t(X))\in(0,1]$ for the corresponding token probabilities.
% The proxy depends on $X$ only through $\{\ell_t(X):t\in\bigcup_{i=1}^T\mathcal{Y}_i\}$. For $t\in\mathcal{Y}_i$, the quantity $\ell_t(X)$ equals the log-probability the model assigns to the true output token at position $t$---recovered for all demonstration positions via a single teacher-forced pass---quantifying the model's predictive confidence on each demonstrated label.

%% ─────────────────────────────────────────────
\paragraph{Component 1 -- Per-demonstration absolute confidence.}
To make confidence comparable across labels with different token lengths, we compute a length-normalized average log-probability for each demonstrated output:
\begin{equation*}
\bar{\ell}_i(X)\;\coloneqq\;\frac{1}{|\mathcal{Y}_i|}\sum_{t\in\mathcal{Y}_i}\ell_t(X).
\end{equation*}
We then map it to a bounded score $c_i(X)\;\coloneqq\;\exp \bigl(\bar{\ell}_i(X)\bigr)\in(0,1]$, which equals the geometric mean of the true-token probabilities over the $i$-th output span. We then average the confidence scores across demonstrations for the absolute confidence, defining
\begin{equation*}
    \bar{C}(X)\coloneqq \frac{1}{T}\sum_{i=1}^T c_i(X).
\end{equation*}

%% ─────────────────────────────────────────────
\paragraph{Component 2 -- Pooled robustness.}
High mean confidence can mask brittleness: a label may look easy on average while containing a few tokens the model assigns very low probability. To penalize such cases, we pool the true-token probabilities $p_t(X)\coloneqq\exp(\ell_t(X))\in(0,1]$ from \emph{all} demonstration output spans into a single set,
\begin{equation*}
\mathcal{S}(X)\;\coloneqq\;\bigl\{p_t(X):t\in\mathcal{Y}_1\cup\mathcal{Y}_2\cup\cdots\cup\mathcal{Y}_T\bigr\},
\end{equation*}
and summarize its lower tail via a fixed quantile level $q\in(0,1)$ (we use $q=0.1$):
\begin{equation*}
R(X)\;\coloneqq\;\operatorname{Quantile}_{q}\!\bigl(\mathcal{S}(X)\bigr)\in[0,1].
\end{equation*}
$R(\cdot)$ is the probability threshold below which the lowest $q$-fraction of all label tokens fall. Viewed as a risk measure, it coincides with Value-at-Risk at level $q$ on the empirical distribution of token confidences, capturing tail fragility that the mean $\bar{C}$ may conceal. Per-demonstration quantiles degenerate for single-token labels ($\operatorname{Quantile}_q(\{p_t\})~=~p_t =~c_i$); pooling across all spans preserves a genuine tail measure.

%% ─────────────────────────────────────────────
\paragraph{Component 3 -- Information gain across demonstrations.}
If the task is inferred progressively, later demonstrations should become more predictable as they are conditioned on a richer in-context history. To capture this trend in a single pass, we track improvements in the per-demonstration confidence sequence $\{c_i(X)\}_{i=1}^T$:
\begin{equation*}
G(X)\;\coloneqq\;
\begin{cases}
\dfrac{1}{T-1}{\displaystyle\sum_{i=2}^{T}}\delta_i(X), & T\ge 2,\\[6pt]
0, & T=1,
\end{cases}
\end{equation*}
where $\delta_i(X)\coloneqq\max\bigl(0,\,c_i(X)-c_{i-1}(X)\bigr)$.
$G(\cdot)$ assigns credit only to increases and remains in $[0,1]$. It is order-dependent by design; rewarding progressive task inference, consistent with evidence that autoregressive models process demonstrations sequentially. Because $G(\cdot)$'s gradient exerts opposing pressure on early- and late-demonstration embeddings under causal attention, we assign it a small weight $\gamma$ so the residual asymmetry remains negligible relative to the confidence and robustness gradients. Conceptually, $G$ can be viewed as a single-pass approximation to the progressive-compression principle underlying description-length approaches~\citep{sa_icl}; rather than evaluating each demonstration in isolation, it tracks confidence improvements as the in-context history grows.

%% ─────────────────────────────────────────────
\paragraph{Final proxy score.}
The final ICL-confidence proxy is the weighted linear combination of the components:
\begin{equation*}
f(X)\;\coloneqq\;\alpha\,\bar{C}(X)+\beta\,R(X)+\gamma\,G(X), \qquad \alpha,\beta,\gamma\ge 0,\;\; \alpha+\beta+\gamma=1.
\end{equation*}
%
% Since each component lies in $[0,1]$, the proxy satisfies~(D1) by construction; it depends exclusively on teacher-forced log-probabilities~(D2); and each component increases as the model's confidence on the demonstrated outputs improves~(D3).
%
Note that we exclude any entropic measure as it is label-agnostic and risks collapse onto incorrect tokens~\citep{come}, and at exemplar positions its signal is already subsumed by $\bar{C}$ and $R$.

%% ─────────────────────────────────────────────
%% Self-Improving In-Context Learning
%% ─────────────────────────────────────────────
\subsection{End-to-End Test-Time Calibration}
\label{sec:final_algorithm}

% We now consolidate the complete iterative procedure.
%
The zeroth-order ascent updates require regularization to prevent embeddings from drifting into out-of-distribution regions of the embedding space. We incorporate three lightweight mechanisms, while deliberately avoiding any external optimization components (e.g., momentum), so that any performance gains can be attributed solely to the test-time embedding adjustment driven by the proposed proxy.

\paragraph{Gradient clipping.}
Clipping preserves proportionality of per-position gradient magnitudes while bounding large updates: $\hat{g}_{k,t}~\leftarrow~\hat{g}_{k,t} / \max(1,\, \|\hat{g}_{k,t}\|_2)$ for each position~$t$.

\paragraph{Cosine similarity.}
We constrain the updated embeddings to remain directionally close to their initial values. At each iteration $k$, if the cosine similarity between the updated embeddings $X_{k+1}$ and the original embeddings $X_0$ falls below a threshold $\kappa \in [0,1]$, we project $X_{k+1}$ back onto the boundary of the feasible region.

\paragraph{Initial proxy gate.}
Before entering the optimization loop, the proxy is evaluated on the unperturbed embeddings. Because each per-exemplar confidence is defined as $c_i = \exp(\bar{\ell}_i)$, its gradient with respect to $\bar{\ell}_i$ equals $c_i$ itself; when $f(X_0)$ is near zero, the proxy surface is exponentially flat and the zeroth-order gradient estimate carries no directional signal. If $f(X_0)$ falls below a threshold $\tau$, optimization is skipped and the original embeddings are returned unchanged.

The complete procedure is summarized in Algorithm~\ref{alg:main} (Appendix~\ref{app:algorithm}). The returned embedding $X^\star$ is the one achieving the highest proxy value across all iterations; after convergence, the query output is generated under $X^\star$ for any desired number of tokens.

\section{Experiments}
\label{sec:experiments}

%% ─────────────────────────────────────────────
%% Experimental Setup
%% ─────────────────────────────────────────────
% \subsection{Experimental Setup}
% \label{sec:experiments_setup}

\paragraph{Benchmark.}
Standard few-shot benchmarks (e.g., MMLU~\citealp{mmlu}) conflate a model's in-context learning ability with its pretraining knowledge and language proficiency~\citep{icleval}: a model may fail not because it cannot learn from demonstrations, but because it lacks the requisite domain knowledge. ICLEval~\citep{icleval} addresses this confound by replacing factual content with hash strings, ensuring that correct predictions can only be derived from the provided demonstrations. The benchmark comprises 12~tasks (2,040~samples) organized around two core sub-abilities: \emph{exact copying} (matching a prefix and reproducing subsequent content) and \emph{rule learning} (inferring format, order, statistical, and list-mapping rules from examples). Each sample includes its own dynamically generated set of 3--31 demonstrations. Further details are provided in Appendix~\ref{sec:experimental_details}.

\paragraph{Baselines.}
We consider inference-time methods of \emph{Contextual Calibration} (CC)~\citep{cc}, which corrects label biases via an affine transformation estimated from content-free inputs; \emph{domain-conditional PMI} (DC-PMI)~\citep{dc_pmi}, which normalizes out the label prior induced by the prompt; and \emph{DEmO}~\citep{demo}, which eliminates ordering sensitivity by processing each exemplar independently and aggregating label-distribution shifts. CC and DC-PMI require a single-token label space and are evaluated on Order Check and Duplication Check (400 of the 2,040 samples). DEmO accommodates multi-token labels and is additionally evaluated on Format Check.

Methods with structural limitations preventing general-purpose use are catalogued in Table~\ref{tab:method_comparison}.

\paragraph{Evaluation.}
Following the ICLEval protocol~\citep{icleval}, we use greedy decoding and report exact-match accuracy---the fraction of samples for which the generated string matches the gold label exactly. Exact match is the natural criterion given the unambiguous, hash-based outputs.

\paragraph{Models.}
We evaluate a range of models spanning different sizes, architectures, and developers: Llama~3.1-8B~\citep{llama3}, Qwen3-4B~\citep{qwen3}, and Gemma~2-2B~\citep{gemma2}. Smaller models such as GPT-2~\citep{gpt2} and Phi-2~\citep{phi} were excluded because their context windows (1024 and 2048 tokens) fall short of the longest prompts in ICLEval (2100 tokens).

\paragraph{Hyperparameters.}
We use $N=16$ for Llama~3.1-8B and $N=8$ for Qwen3-4B and Gemma~2-2B. The configuration $\alpha = 0.6$, $\beta = 0.3$, $\gamma = 0.1$ is selected; the small value of $\gamma$ keeps the information-gain term from suppressing absolute confidence (Component~1). The proxy gate threshold is set to $\tau=0.05$. Optimization is terminated when the proxy fails to improve for 5 consecutive steps. Complete hyperparameter settings and sweep details are provided in Appendix~\ref{app:hyperparams}.

%% ─────────────────────────────────────────────
%% Main Results
%% ─────────────────────────────────────────────

\begin{table*}[t]
  \centering
  \small
  \begin{tabular}{l r ccr c ccr c ccr}
      \toprule
      & & \multicolumn{3}{c}{Llama~3.1-8B} & & \multicolumn{3}{c}{Qwen3-4B} & & \multicolumn{3}{c}{Gemma~2-2B} \\
      \cmidrule(lr){3-5} \cmidrule(lr){7-9} \cmidrule(lr){11-13}
      Task & $n$ & Base & Ours & $\Delta$ (\%) & & Base & Ours & $\Delta$ (\%) & & Base & Ours & $\Delta$ (\%) \\
      \midrule
      String Completion    & 100 & 0.57 & 0.57          & $=$       && 0.90 & 0.90          & $=$       && 0.87 & 0.87          & $=$       \\
      Dict.\ Search        & 190 & 0.87 & 0.87          & $=$       && 0.92 & 0.92          & $=$       && 0.66 & 0.66          & $=$       \\
      \addlinespace[0.7em]
      Format Check         & 120 & 0.07 & \best{0.17} & $+122.2$  && 0.17 & \best{0.36} & $+104.8$  && 0.07 & 0.07          & $=$       \\
      Format Cloning       & 100 & 0.97 & 0.97          & $=$       && 0.85 & \best{0.89} & $+4.7$    && 0.71 & \best{0.84} & $+18.3$   \\
      Format Conversion    & 120 & 0.86 & \best{0.88} & $+2.9$    && 0.72 & \best{0.77} & $+5.7$    && 0.65 & \best{0.71} & $+9.0$    \\
      \addlinespace[0.7em]
      Order Check          & 100 & 0.98 & 0.98          & $=$       && 1.00 & 1.00          & $=$       && 0.78 & \best{0.79} & $+1.3$    \\
      Order Adjustment     & 240 & 0.86 & \best{0.95} & $+10.7$   && 0.60 & \best{0.70} & $+18.2$   && 0.37 & \best{0.40} & $+6.7$    \\
      \addlinespace[0.7em]
      Duplication Check    & 300 & 0.69 & \best{0.76} & $+11.7$   && 0.73 & \best{0.75} & $+3.2$    && 0.49 & \best{0.53} & $+7.4$    \\
      De-Duplication       & 300 & 0.75 & \best{0.85} & $+13.8$   && 0.70 & \best{0.85} & $+20.4$   && 0.24 & \best{0.28} & $+15.1$   \\
      Count \& Navigation  & 120 & 0.29 & \best{0.34} & $+17.1$   && 0.43 & \best{0.51} & $+17.3$   && 0.03 & 0.03          & $=$       \\
      Relation Analysis    & 100 & 0.47 & \best{0.56} & $+19.2$   && 0.27 & \best{0.35} & $+29.6$   && 0.03 & \best{0.12} & $+300.0$  \\
      \addlinespace[0.7em]
      List Mapping         & 250 & 0.63 & \best{0.66} & $+5.1$    && 0.56 & \best{0.62} & $+11.4$   && 0.48 & 0.48          & $=$       \\
      \midrule
      Mean                 &     & 0.67 & \best{0.71} & $+6.0$    && 0.65 & \best{0.72} & $+10.8$   && 0.45 & \best{0.48} & $+6.7$    \\
      \bottomrule
  \end{tabular}
  \caption{Per-task exact-match accuracy on ICLEval ($n$: number of test samples). Boldface indicates strict improvement over the base model; $\Delta$ denotes relative improvement (\%). One-sided McNemar test ($H_1$: Ours $>$ Base) across all 2{,}040 samples: $p = 0.001$ for Llama~3.1-8B, $p = 0.038$ for Qwen3-4B, $p < 0.001$ for Gemma~2-2B. Baseline comparisons on the classification subset are reported in Table~\ref{tab:baselines}.}
  \label{tab:results}
  \end{table*}

\subsection{Main Results}

Table~\ref{tab:results} reports per-task results for all three models. Across all three models, our method either matches or improves base-model accuracy on every task, never causing degradation---a property enforced by the regularization mechanisms. The per-model improvements are statistically significant under the one-sided McNemar test ($p = 0.001$ for Llama, $p = 0.038$ for Qwen, $p < 0.001$ for Gemma), which operates at the individual-sample level and accounts for both improved and degraded predictions.

On exact-copying tasks (String Completion, Dictionary Search), accuracy remains unchanged across all models. Here, improvements in the proxy do not transfer to query-level accuracy, i.e., increasing confidence on exemplar sequences does not help the model reproduce a specific target hash sequence.

For rule-learning tasks, improvement depends not on base accuracy alone but on whether the model possesses a latent task-specific capability that the proxy can surface. Gemma~2-2B illustrates this clearly: Relation Analysis and Count~\&~Navigation share the same base accuracy (0.03), yet only the former improves ($0.03 \to 0.12$). At the opposite extreme, near-ceiling tasks leave no room for further gains. Between these two regimes, where capability exists but remains underutilized, improvements are consistent across format, order, statistics, and list-mapping rules.

\paragraph{Comparison with the baselines.}
Table~\ref{tab:baselines} reports the numerical comparison on the classification subset. Our method outperforms all baselines on Llama and Qwen despite operating without a predefined label space. On Gemma---the smallest and most ordering-sensitive model---DEmO is the strongest baseline, most notably on Format Check ($0.85$ vs.\ $0.07$): by processing exemplars independently, DEmO bypasses the compositional difficulty Gemma faces when attending over the full demonstration sequence. On larger models, DEmO's ordering-based approach yields diminishing returns. CC and DC-PMI degrade Duplication Check accuracy on Llama and Qwen through overcalibration of their content-free bias estimates. All baselines require a finite label space, restricting them to the classification subset, whereas the proposed method applies fully task-agnostic.

\begin{table*}[t]
  \centering
  \small
  \begin{tabular*}{\linewidth}{@{\extracolsep{\fill}} l l cccc @{}}
      \toprule
      Task & Model & CC & DC-PMI & DEmO & Ours \\
      \midrule
      \multirow{3}{*}{Format Check}
          & Llama~3.1-8B & \textemdash & \textemdash & 0.16                                    & \highlight[customorange]{\best{0.17}} \\
          & Qwen3-4B     & \textemdash & \textemdash & 0.17                                    & \highlight[customorange]{\best{0.36}} \\
          & Gemma~2-2B   & \textemdash & \textemdash & \highlight[customorange]{\best{0.85}}  & 0.07                                   \\
      \addlinespace[0.75em]
      \multirow{3}{*}{Order Check}
          & Llama~3.1-8B & \highlight[customblue]{0.98} & \highlight[customblue]{0.98} & \highlight[customblue]{0.98}            & \highlight[customblue]{0.98}            \\
          & Qwen3-4B     & \highlight[customblue]{1.00} & \highlight[customblue]{1.00} & \highlight[customblue]{1.00}            & \highlight[customblue]{1.00}            \\
          & Gemma~2-2B   & 0.79                         & 0.70                         & \highlight[customorange]{\best{0.87}} & 0.79                                   \\
      \addlinespace[0.75em]
      \multirow{3}{*}{Duplication Check}
          & Llama~3.1-8B & 0.60 & 0.67 & 0.70                                    & \highlight[customorange]{\best{0.76}} \\
          & Qwen3-4B     & 0.60 & 0.73 & 0.68                                    & \highlight[customorange]{\best{0.75}} \\
          & Gemma~2-2B   & 0.48 & 0.50 & \highlight[customorange]{\best{0.56}}  & 0.53                                   \\
      \bottomrule
  \end{tabular*}
  \caption{Baseline comparison on the classification subset of ICLEval. CC and DC-PMI require a single-token label space (Order Check and Duplication Check only); DEmO also applies to Format Check. \highlight[customorange]{\best{Best}} and \highlight[customblue]{tied best} results are highlighted.}
  \label{tab:baselines}
\end{table*}

\paragraph{Computational cost.}
Each optimization step requires $N{+}1$ forward passes (one base evaluation and $N$ perturbations). Because the proxy is non-stationary (i.e., each embedding update changes the log-probabilities from which it is computed), conservative learning rates are necessary; per-task iteration counts are reported in Table~\ref{tab:iteration_counts} (Appendix~\ref{app:iterations}).

\subsection{Proxy--Performance Correlation}

\begin{figure*}[t]
\centering
\subfigure[Llama 3.1-8B]{%
  \includegraphics[width=0.30\textwidth]{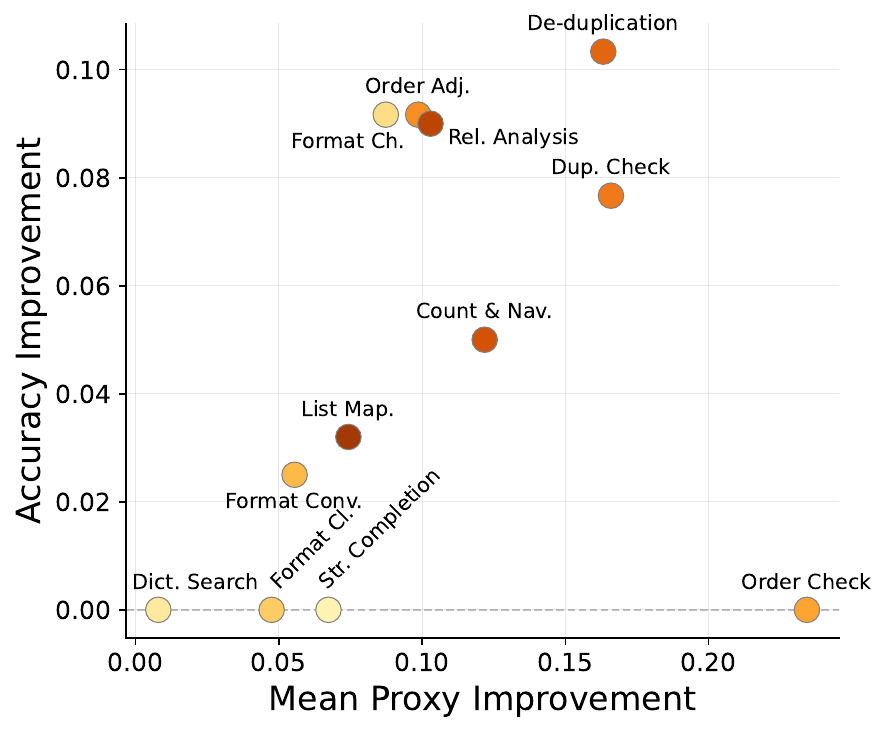}%
  \label{fig:proxy_acc_llama}%
}
\hfill
\subfigure[Qwen3-4B]{%
  \includegraphics[width=0.30\textwidth]{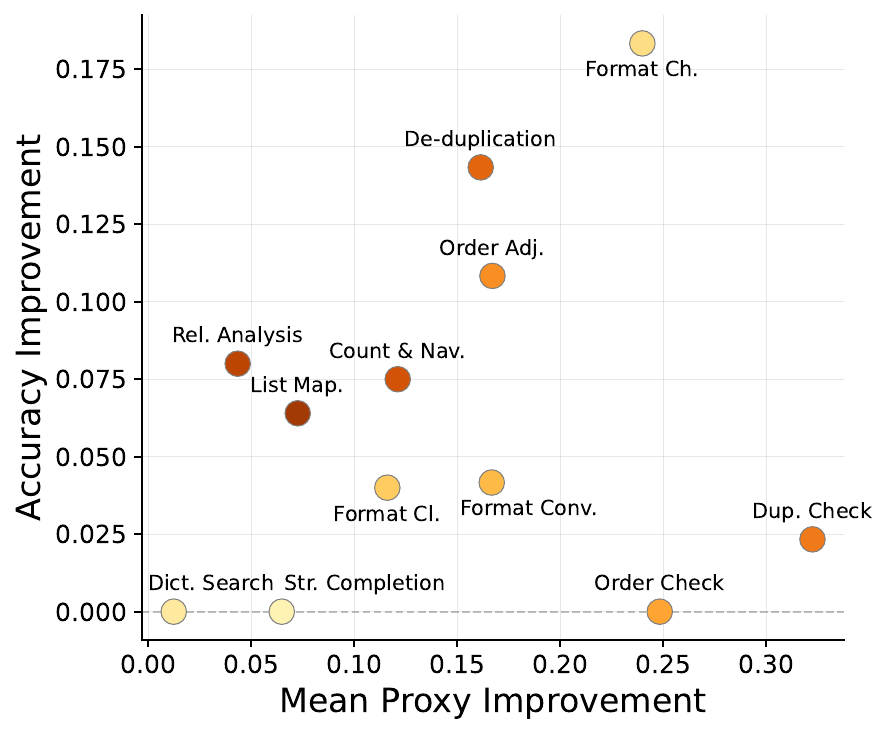}%
  \label{fig:proxy_acc_qwen}%
}
\hfill
\subfigure[Gemma~2-2B]{%
  \includegraphics[width=0.30\textwidth]{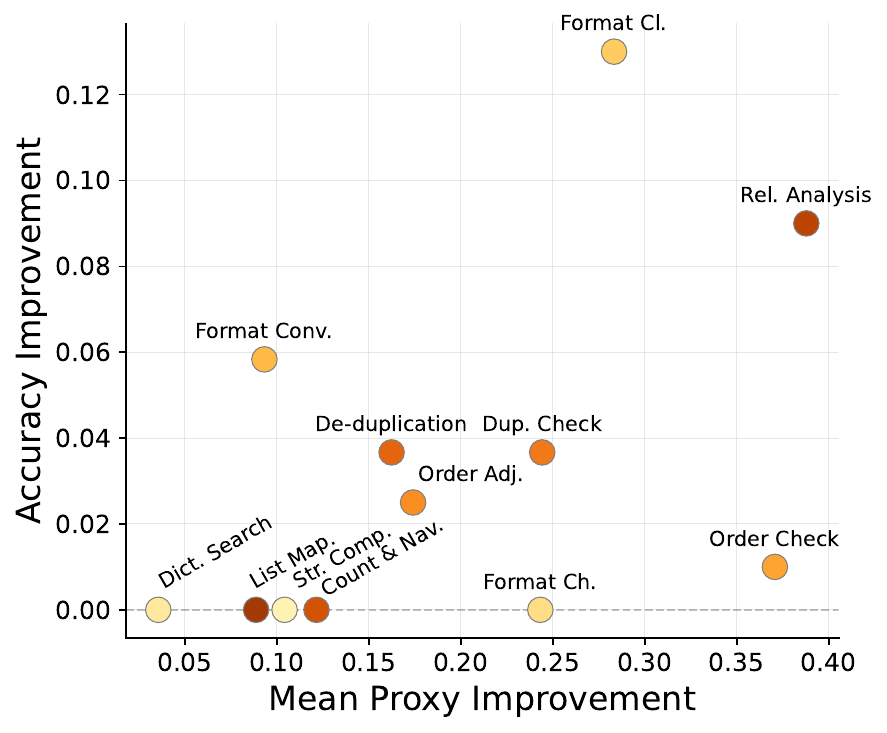}%
  \label{fig:proxy_acc_gemma}%
}
\caption{Proxy improvement versus accuracy improvement across all 12 ICLEval tasks. A one-sided Spearman rank correlation test ($H_1\colon \rho > 0$) yields a statistically significant positive association across all models and tasks combined.}
\label{fig:proxy_vs_accuracy}
\end{figure*}

For the proxy to serve as a meaningful optimization signal, improvements in its value should translate to improvements in downstream performance. We test this by examining whether tasks with greater proxy improvement also exhibit greater accuracy gains. For each of the 12 tasks, we compute the mean per-sample proxy improvement and the change in accuracy between the optimized and baseline prompts. Figure~\ref{fig:proxy_vs_accuracy} plots accuracy improvement against proxy improvement, with each point representing one task. To quantify the monotonic association, we apply a one-sided Spearman rank correlation test ($H_1\colon \rho > 0$). Across all 12 tasks, the correlation between proxy and accuracy improvements is statistically significant for Llama and Gemma, but not for Qwen.

Two tasks account for the weaker correlation observed in Qwen. Order Check achieves perfect baseline accuracy, leaving no room for improvement and rendering its inclusion uninformative. Duplication Check exhibits large proxy improvement with minimal accuracy gain: its label vocabulary consists of only two tokens (True/False), so the task is already well understood by the model. Continued optimization beyond consolidating this understanding begins to overfit to the output format rather than improving the input--output mapping, inflating confidence without improving discrimination. These proxy gains do not transfer to the query position; the model becomes more certain about predicting True or False, but no more accurate. We identify this as a mild limitation as it can be addressed with stricter early stopping.

Excluding these two tasks, the Spearman test yields a statistically significant positive rank correlation at 95\% confidence. Fisher's combined Spearman test across all models achieves statistical significance ($\chi^2 = 13.95$, $p = 0.03$) even when Order Check and Duplication Check are included, confirming that the \textit{ICL confidence proxy is a meaningful optimization signal for in-context learning}.

\begin{figure*}[t]
  \centering
      \subfigure[Proxy component ablation]{%
        \includegraphics[height=5.5cm]{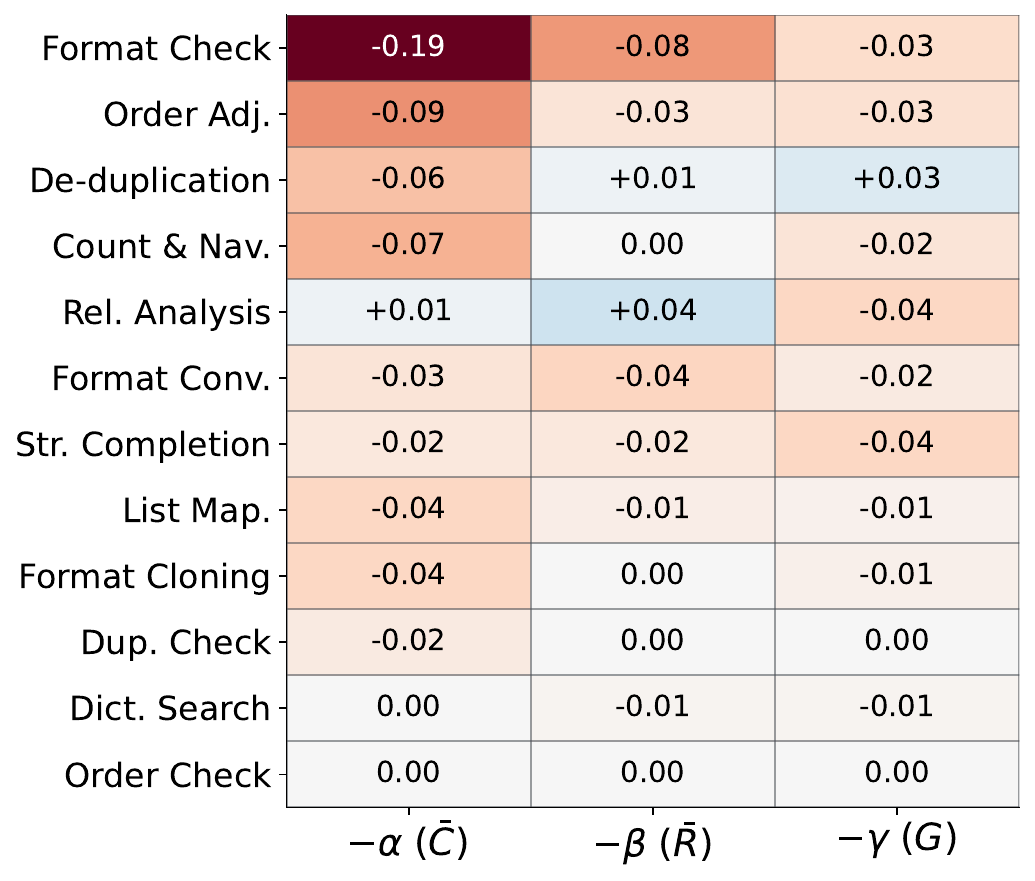}%
        \label{fig:ablation_components}%
      }
      \hfill
      \subfigure[Perturbation domain]{%
        \includegraphics[height=5.5cm]{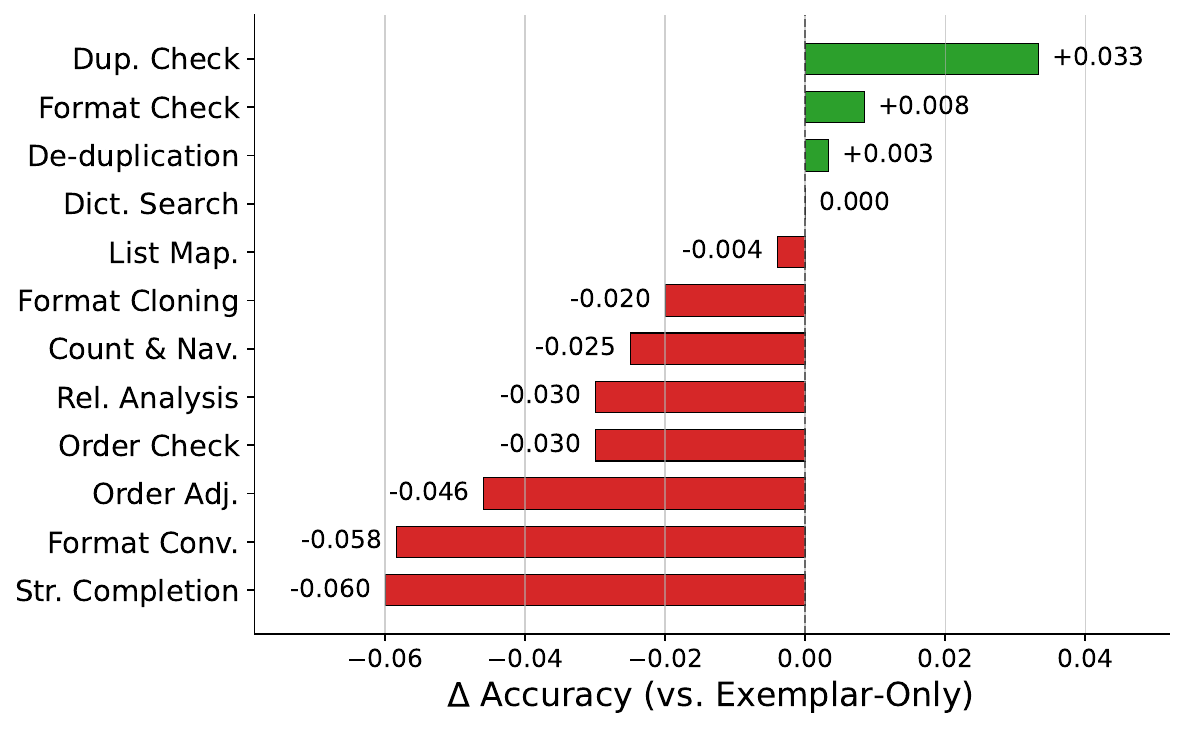}%
        \label{fig:ablation_perturbation}%
      }
      \caption{Ablation studies of the proposed three-component proxy ($\alpha{=}0.6$, $\beta{=}0.3$, $\gamma{=}0.1$) and the perturbation domain used in end-to-end calibration. (a)~Each proxy component is ablated by setting its coefficient to zero and renormalizing the remaining weights. (b)~Accuracy difference between the default perturbation domain (demonstration embeddings only) and full-sequence perturbation that includes query positions. Results are reported for Qwen3-4B.}
  \label{fig:ablations}
\end{figure*}

\subsection{Ablations and Sensitivity Analysis}

To understand the contribution of each proxy component, we perform ablations by setting the corresponding coefficient to zero and renormalizing the remaining weights relative to the original configuration ($\alpha{=}0.6$, $\beta{=}0.3$, $\gamma{=}0.1$). We also examine the effect of perturbing the full token sequence (including query positions) versus perturbing only the demonstration embeddings, as in the default setting. The results are shown in Figure~\ref{fig:ablations}.

\begin{figure*}[!t]
  \centering
      \subfigure[$N$ (Monte Carlo samples)]{%
        \includegraphics[height=8.25cm]{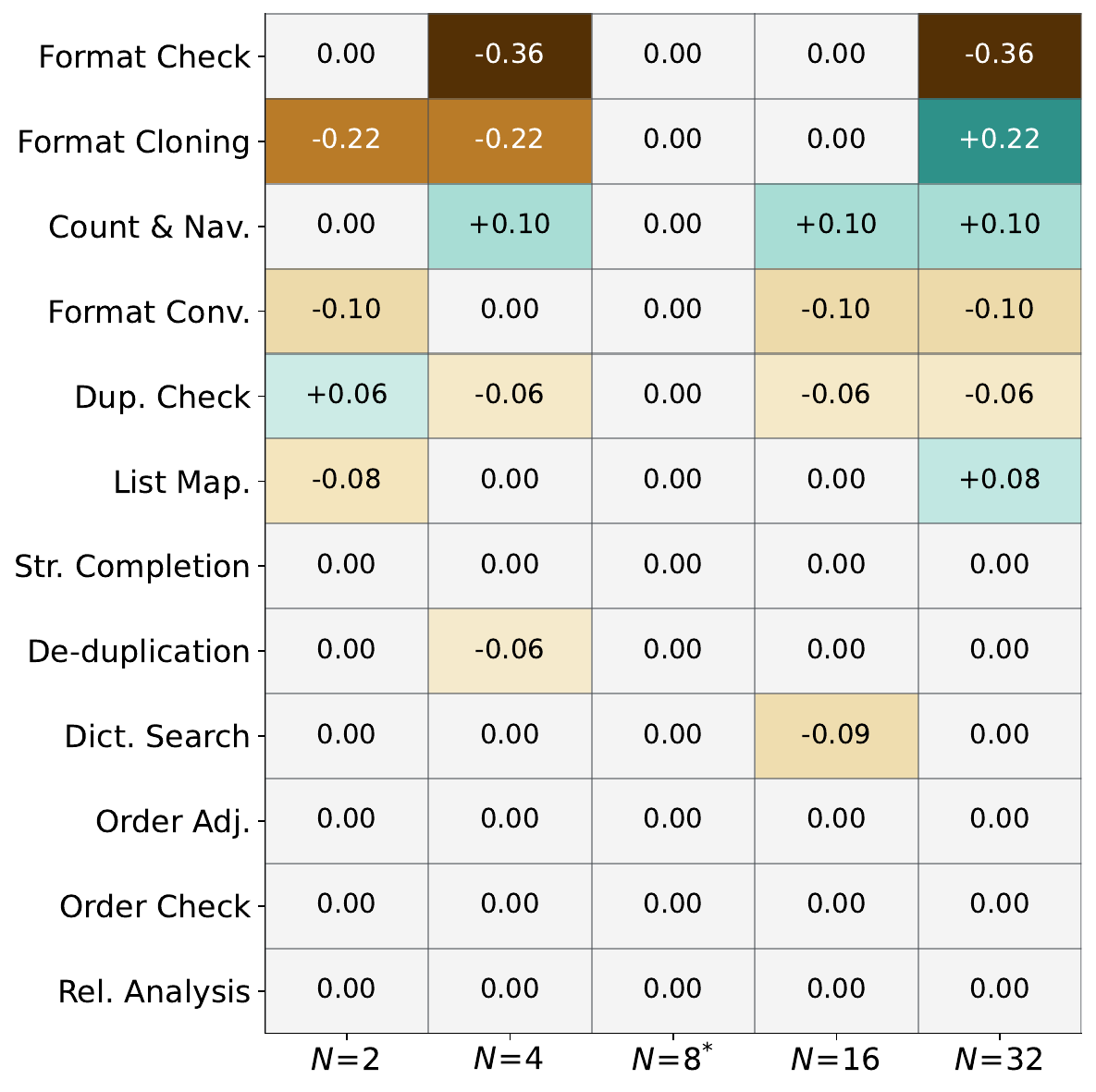}%
        \label{fig:sensitivity_N}%
      }
      \hfill
      \subfigure[$\mu$ (perturbation scale)]{%
        \includegraphics[height=8.25cm]{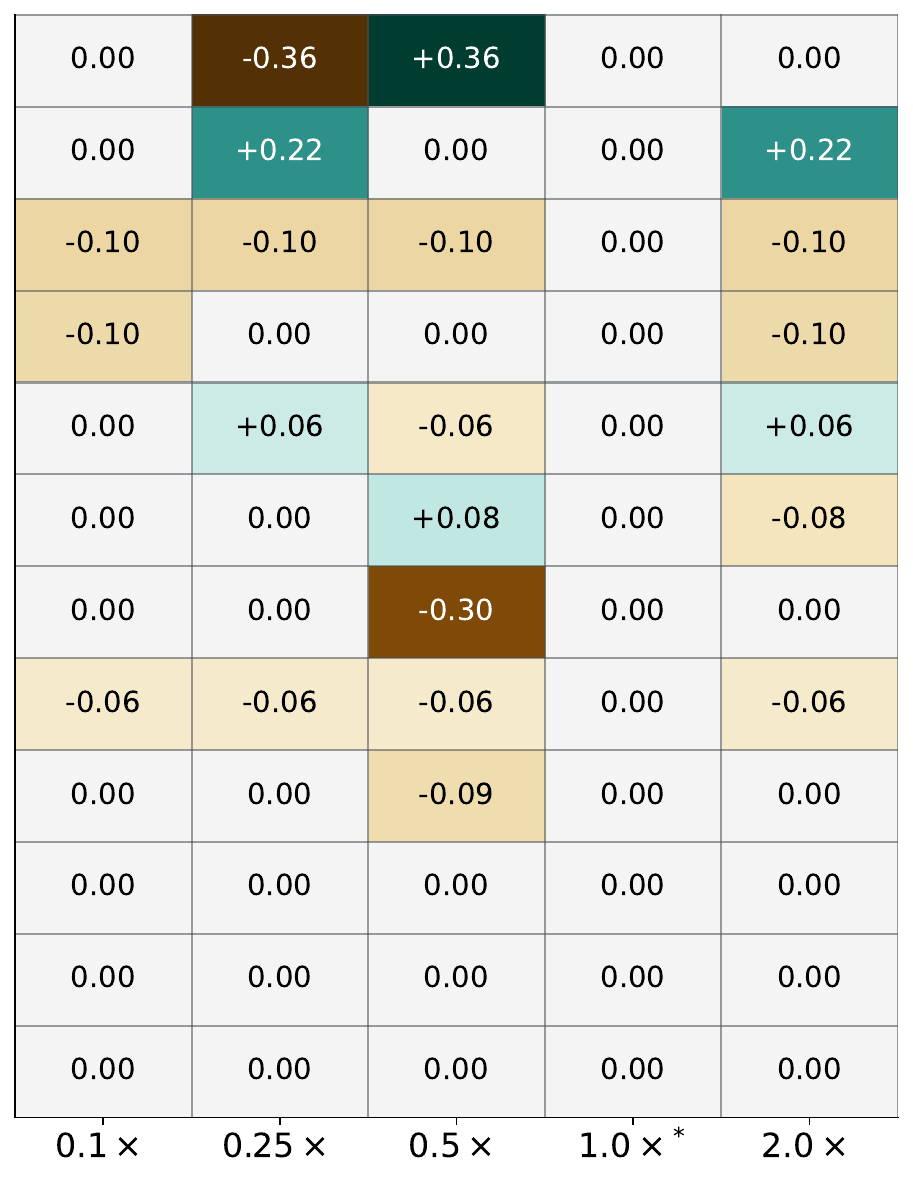}%
        \label{fig:sensitivity_mu}%
      }
      \caption{Downstream accuracy under varying perturbation sample counts $N$ and perturbation scales, expressed as fractions of the optimal value $\mu = 0.004$, with all other hyperparameters held fixed. Results are reported for Qwen3-4B, where $(^*)$ indicates the values used in the main experiments.}
  \label{fig:sensitivity}
  \end{figure*}

The confidence component ($\bar{C}$) is clearly the primary driver of the method's gains: removing it reduces the fraction of samples that improve during optimization from over 90\% to roughly half, and cuts the average proxy improvement by approximately 40\%. The robustness component ($R$) follows in importance---without it, the optimization remains broadly effective but achieves lower proxy improvement, indicating that low-probability outlier tokens are common enough for the tail-confidence signal to provide a meaningful complementary gradient direction that the mean alone cannot capture. Removing the information gain component ($G$) has the least effect; its small weight ($\gamma = 0.1$) means the redistribution to the remaining two components preserves---and even marginally sharpens---the optimization signal. Perturbing the full token sequence degrades downstream accuracy on the tasks whose optimization trajectory is affected, confirming that noise at query positions, where the proxy gradient has zero expectation, accumulates into drift that distorts the query representation without a compensating proxy benefit.

The quality of the zeroth-order gradient estimates depends directly on the perturbation strength $\mu$ and the number of perturbation samples $N$, the latter of which also determines computational cost. We sweep several values of each around the optimal configuration; results are shown in Figure~\ref{fig:sensitivity}. Downstream accuracy varies non-monotonically with both parameters, consistent with the standard bias--variance tradeoff in finite-difference gradient estimation (Section~\ref{sec:zo}).
\section{Conclusion}
\label{sec:conclusion}

We presented a test-time method that improves in-context learning through zeroth-order optimization of the prompt's continuous embeddings, without finetuning the model, generating any tokens, or requiring auxiliary data. The optimization objective is a bounded, self-supervised confidence proxy derived entirely from the model's log-probabilities on its demonstrated outputs, capturing per-demonstration predictive confidence, tail robustness across output tokens, and progressive improvement in predictions along the demonstration sequence. Across models spanning 2B to 8B parameters, the method never degrades base-model accuracy and with statistically significant improvements across all evaluated models, it outperforms classification-specific baselines on most tasks despite requiring no predefined label set.

The statistically significant correlation between proxy and accuracy improvement, consistent across diverse task types and model scales, confirms that demonstration log-probabilities constitute a reliable optimization surface---one whose ascent direction aligns with downstream performance, rather than merely a scoring function for selecting or ranking prompts. A natural extension is to compose this calibration with existing demonstration selection or ordering strategies---for instance, applying embedding optimization on a prompt whose demonstrations have already been selected or reordered.

\section*{Limitations}
The method requires that output-span positions are known and that exemplar labels are correct; it has not been validated on prompts with fewer than three demonstrations. Additionally, since the proxy is non-stationary (each embedding update changes the log-probabilities from which it is computed), conservative learning rates are necessary, which may increase the number of optimization steps. In practice, however, this overhead remained modest: the longest per-sample runtime in our experiments (Qwen3-4B on Format Check; see Table~\ref{tab:iteration_counts} in the appendix) was roughly a minute and could be further reduced with optimized inference stacks (see Appendix~\ref{app:implementation}).

%%%%%%%%%%%%%%%%%%%%%%%%%%%%%%%%%%%%%%%%%%%%%%%%%%%%%%%%%%%%%%%%%%%%%%%%%%%%%%%%
% ACKNOWLEDGEMENTS
%%%%%%%%%%%%%%%%%%%%%%%%%%%%%%%%%%%%%%%%%%%%%%%%%%%%%%%%%%%%%%%%%%%%%%%%%%%%%%%%
% \section*{Acknowledgements}
%
% Acknowledgements go here.

%%%%%%%%%%%%%%%%%%%%%%%%%%%%%%%%%%%%%%%%%%%%%%%%%%%%%%%%%%%%%%%%%%%%%%%%%%%%%%%%
% REFERENCES
%%%%%%%%%%%%%%%%%%%%%%%%%%%%%%%%%%%%%%%%%%%%%%%%%%%%%%%%%%%%%%%%%%%%%%%%%%%%%%%%
\bibliographystyle{plainnat}
\bibliography{references}

%%%%%%%%%%%%%%%%%%%%%%%%%%%%%%%%%%%%%%%%%%%%%%%%%%%%%%%%%%%%%%%%%%%%%%%%%%%%%%%%
% APPENDIX
%%%%%%%%%%%%%%%%%%%%%%%%%%%%%%%%%%%%%%%%%%%%%%%%%%%%%%%%%%%%%%%%%%%%%%%%%%%%%%%%
\clearpage
\appendix

\section{Applicability of Existing Test-Time Methods}
\label{sec:excluded_baselines}

Examined in Section~\ref{sec:related_work}, existing methods span over demonstration selection, demonstration ordering, and output calibration. While effective in their respective target settings---predominantly text classification with candidate demonstration pools---each category carries structural assumptions that prevent broad applicability. This discussion explains why these methods cannot serve as general-purpose baselines in our experiments, and why the three baselines we do report (CC, DC-PMI, DEmO) are restricted to the classification subset.

We identify three failure modes: \emph{(i)} restriction to classification tasks with a finite label space, \emph{(ii)} dependence on the latent semantic structure of prompt inputs, and \emph{(iii)} combinatorial scaling in the number of demonstrations. These are inherent properties of the methods, not artifacts of any particular benchmark; ICLEval makes all three visible in a single evaluation because it combines classification and generation tasks, masks factual content with hash strings, and the number of in-context examples can be as many as 31.

Table~\ref{tab:method_comparison} provides a systematic comparison along four operational axes that expose these structural limitations.

\subsection{Demonstration Selection}

Selection methods choose which examples to place in the prompt from an external candidate pool. Their gains depend on the quality of this choice, but the mechanism by which candidates are scored determines whether those gains reflect genuine ICL improvement or exploitation of the model's parametric knowledge.

\paragraph{Knowledge dependence.}
Similarity-based methods~\citep{kate, vote_k, dpp_icl} retrieve candidates whose inputs are semantically or lexically close to the test query. This strategy succeeds when some pretrained representation encodes task-relevant structure that transfers via surface similarity. When this structure is absent---for example, when factual tokens are replaced with opaque identifiers such as hash strings in ICLEval---the features these methods match on become uninformative, and selection reduces to random. This reveals that similarity-based selection improves ICL performance partly by routing the model toward inputs where its parametric knowledge is already useful, rather than by strengthening the ICL mechanism itself.

LLM-feedback methods~\citep{sa_icl, cone, se2, lens_icl, misconfidence, ids_icl, dva_icl} avoid this dependence by scoring candidates using the model's own output distributions, while task-specific heuristics select for reasoning complexity~\citep{complexity_prompting} or structural coverage~\citep{diverse_demos}. In both cases, the scoring criteria are typically tied to a finite label space or a specific task format and do not extend to open-ended generation.

\paragraph{Candidate pool requirement.} All selection methods presuppose a pool of candidate demonstrations external to the prompt. Constructing one---whether by domain-specific curation or by aggregating exemplars across test instances---introduces assumptions about data availability that fall outside the scope of test-time adaptation. Self-generated methods~\citep{sg_icl, z_icl, usp, cosp} bypass the pool requirement by generating demonstrations from scratch, but they replace the exemplar content entirely rather than improving ICL over the given original prompt. They may also require explicit task descriptions or an external text corpus, neither of which is available in plug-and-play settings.

\subsection{Demonstration Ordering}

Ordering methods optimize the sequence in which exemplars appear in the prompt. The well-documented sensitivity of ICL to ordering~\citep{lu_ordering, demo} motivates a body of work that searches over permutations to find the best arrangement.

\paragraph{Task generality.} The scoring functions used to evaluate candidate orderings (e.g., entropy over predicted label distributions~\citealp[GlobalE/LocalE]{lu_ordering}, label fairness~\citealp{demo}, label-distribution optimization~\citealp{xu_label_dist_ordering}, prompt embedding clustering~\citealp{cluster_search}) all require a finite, enumerable label set. For open-ended generation tasks, there is no such set, and these scoring functions are undefined. Batch-ICL~\citep{batch_icl} takes a different approach by eliminating order sensitivity entirely: it processes each exemplar as an independent one-shot prompt and aggregates the resulting output-distribution shifts onto a zero-shot query. The aggregation, however, operates over label probabilities and has been evaluated exclusively on classification benchmarks. In ICLEval, 74.5\% of the benchmark (1,520 of 2,040 samples) consists of generation tasks with free-form outputs, making these methods inapplicable to the majority of the evaluation.

\paragraph{Scalability.} OEOICL~\citep{oeoicl} is the only ordering method whose scoring function---log-probability distinguishability of the generated output---does not require a label space and could in principle extend to generation tasks. However, it evaluates all $M!$ permutations of the demonstrations. For $M = 8$ this requires approximately 40,000 forward passes per test instance; for $M \geq 10$ it is computationally intractable. Even with sampling, the cost grows combinatorially with the number of demonstrations---a limitation shared by any method that searches the permutation space. ICLEval includes tasks with up to 31 in-context examples, placing a large portion of the benchmark well beyond the reach of permutation-based methods.

\subsection{Output Calibration}

Calibration methods adjust the model's output probabilities to correct systematic biases---majority-label bias, recency bias, common-token bias, and surface-form competition.

\paragraph{Task generality.}
Calibration methods usually operate by scoring, rescoring, or comparing probabilities across a known set of candidate labels~\citep{cc, dc_pmi, noisy_channel, batch_cal, answer-level-calibration, knn_prompting}. In every case, the method requires a finite label set over which to operate. When the output is free-form text---as in format conversion, deduplication, sequence completion---there is no label set to calibrate over. Additionally, several calibration methods impose further constraints, such as requiring transductive access to a batch of test inputs~\citep{batch_cal, proca, pride} and to the model's hidden states~\citep{hidden_cal}, or assuming an NLI-style premise--hypothesis input structure~\citep{task_cal}.

The proposed method does not exhibit any of the above limitations. It accepts a fixed input prompt without selecting or reordering demonstrations, operates identically on classification and open-ended generation tasks, and its computational cost is determined by a fixed perturbation batch size that does not scale with the number of in-context examples.

\begin{table*}[htpb]
    \centering
    \begin{threeparttable}[b]
    \footnotesize
        \begin{tabular*}{\linewidth}{@{\extracolsep{\fill}} c l cccc @{}}
            \toprule
            & Method & Test-time & No hidden repr. & Task-agnostic & Self-contained \\
            \midrule
            \multirow{17}{*}{\rotatebox[origin=c]{90}{Selection}}
            & Complexity-Based~\citep{complexity_prompting} & \cmark & \cmark & \xmark & \xmark \\
            & COSP\tnote{$\dagger$}~\citep{cosp}            & \cmark & \cmark & \xmark & \cmark \\
            & D.Va~\citep{dva_icl}                          & \cmark & \cmark & \cmark & \xmark \\
            & Diverse Demos~\citep{diverse_demos}           & \cmark & \cmark & \xmark & \xmark \\
            & IDS~\citep{ids_icl}                           & \cmark & \cmark & \xmark & \xmark \\
            & KATE~\citep{kate}                             & \cmark & \cmark & \cmark & \xmark \\
            & LENS~\citep{lens_icl}                         & \cmark & \cmark & \xmark & \xmark \\
            & Misconfidence~\citep{misconfidence}           & \cmark & \cmark & \xmark & \xmark \\
            & Se$^2$~\citep{se2}                            & \cmark & \cmark & \xmark & \xmark \\
            & Self-Adaptive (MDL)~\citep{sa_icl}            & \cmark & \cmark & \xmark & \xmark \\
            & SG-ICL\tnote{$\dagger$}~\citep{sg_icl}        & \cmark & \cmark & \xmark & \cmark \\
            & TopK+ConE~\citep{cone}                        & \cmark & \cmark & \xmark & \xmark \\
            & Two-Stage DPP~\citep{dpp_icl}                 & \cmark & \cmark & \cmark & \xmark \\
            & USP\tnote{$\dagger$}~\citep{usp}              & \cmark & \cmark & \cmark & \xmark \\
            & Vote-k~\citep{vote_k}                         & \cmark & \cmark & \cmark & \xmark \\
            & Z-ICL~\citep{z_icl}                           & \cmark & \cmark & \xmark & \xmark \\
            \midrule
            \multirow{6}{*}{\rotatebox[origin=c]{90}{Ordering}}
            & Batch-ICL\tnote{$\ddagger$}~\citep{batch_icl}        & \cmark & \xmark & \xmark & \cmark \\
            & Cluster-Based Search~\citep{cluster_search}          & \cmark & \xmark & \xmark & \cmark \\
            & DEmO~\citep{demo}                                    & \cmark & \cmark & \xmark & \cmark \\
            & GlobalE / LocalE~\citep{lu_ordering}                 & \cmark & \cmark & \xmark & \cmark \\
            & Label Dist.\ Ordering~\citep{xu_label_dist_ordering} & \cmark & \cmark & \xmark & \cmark \\
            & OEOICL\tnote{$\star$}~\citep{oeoicl}                 & \cmark & \cmark & \cmark & \cmark \\
            \midrule
            \multirow{12}{*}{\rotatebox[origin=c]{90}{Calibration}}
            & Answer-Level~\citep{answer-level-calibration}        & \cmark & \cmark & \xmark & \cmark \\
            & Batch Calibration~\citep{batch_cal}                  & \cmark & \cmark & \xmark & \xmark \\
            & CC~\citep{cc}                                        & \cmark & \cmark & \xmark & \cmark \\
            & DC-PMI~\citep{dc_pmi}                                & \cmark & \cmark & \xmark & \cmark \\
            & Domain-Context~\citep{dc_cal}                        & \cmark & \cmark & \xmark & \xmark \\
            & Hidden Calibration~\citep{hidden_cal}                & \cmark & \xmark & \xmark & \xmark \\
            & kNN Prompting~\citep{knn_prompting}                  & \cmark & \cmark & \xmark & \xmark \\
            & Noisy Channel~\citep{noisy_channel}                  & \cmark & \cmark & \xmark & \cmark \\
            & PriDe~\citep{pride}                                  & \cmark & \cmark & \xmark & \xmark \\
            & ProCa~\citep{proca}                                  & \cmark & \cmark & \xmark & \xmark \\
            & Task Calibration~\citep{task_cal}                    & \cmark & \cmark & \xmark & \cmark \\
            \midrule
            & \textbf{Ours}                                        & \cmark & \cmark & \cmark & \cmark \\
            \bottomrule
        \end{tabular*}
        \vspace{6pt}
        \begin{tablenotes}\footnotesize
            \item[\textit{$\dagger$}] Generates demonstrations from scratch, replacing the given prompt rather than improving it.
            \item[\textit{$\ddagger$}] An output-level variant that aggregates label-probability shifts without hidden representation access is also conceivable; however, this approach substantially underperformed the base model in our initial experiments.
            \item[\textit{$\star$}] Evaluates all $M!$ orderings; intractable for large $M$.
            \end{tablenotes}
    \end{threeparttable}
    \caption{Operational comparison of prior methods against the proposed approach. \cmark{} and \xmark{} indicate whether the property is satisfied or not. \underline{Test-time}: no training or optimization of any module over a dataset. \underline{No hidden representations}: no access to the model's hidden representations (hidden states, attention weights, or gradients); access to output log-probabilities and input word embeddings is permitted. \underline{Task-agnostic}: applicable to general-purpose settings; not restricted to classification, structured reasoning, or other task-specific formats. \underline{Self-contained}: requires no external data such as candidate pools, test batches, or corpora; methods that construct reference inputs or demonstrations internally satisfy this criterion.}
    \label{tab:method_comparison}
\end{table*}

%%%%%%%%%%%%%%%%%%%%%%%%%%%%%%%%%%%%%%%%%%%%%%%%%%%%%%%%%%%%%%%%%%%%%%%%%%%%

\begin{table*}[htpb]
  \centering
  \small
      \begin{tabular}{llcccccc}
      \toprule
      & & & & \multicolumn{3}{c}{\textbf{\# Tokens}} & \\
      \cmidrule(lr){5-7}
      \textbf{Category} & \textbf{Task} & \textbf{\# Samples} & \textbf{\# Demos} & \textbf{Min} & \textbf{Max} & \textbf{Avg} & \textbf{Type} \\
      \midrule
      \multirow{3}{*}{Exact Copying}
        & String Completion     & 100  & \textemdash & 246   & 2,100 & 1,045 & Generation \\
        & Dict.\ Search (String) & 100  & 19  & 468   & 724   & 579   & Generation \\
        & Dict.\ Search (Number) & 90   & 10  & 1,064 & 1,084 & 1,074 & Generation \\
      \midrule
      \multirow{3}{*}{Format Rules}
        & Format Check          & 120  & 6   & 192   & 230   & 207   & Clf. (multi-token) \\
        & Format Cloning        & 100  & 5   & 429   & 1,197 & 655   & Generation \\
        & Format Conversion     & 120  & 3   & 137   & 1,592 & 478   & Generation \\
      \midrule
      \multirow{2}{*}{Order Rules}
        & Order Check           & 100  & 8   & 295   & 319   & 302   & Clf. (single-token) \\
        & Order Adjustment      & 240  & 5   & 134   & 1,157 & 395   & Generation \\
      \midrule
      \multirow{4}{*}{Statistics Rules}
        & Duplication Check     & 300  & 8   & 124   & 1,362 & 475   & Clf. (single-token) \\
        & De-Duplication        & 300  & 5   & 198   & 1,469 & 536   & Generation \\
        & Count \& Navigation   & 120  & 8   & 127   & 389   & 236   & Generation \\
        & Relation Analysis     & 100  & 5   & 542   & 1,043 & 686   & Generation \\
      \midrule
      List Mapping
        & Numbers' List Mapping & 250  & 31  & 426   & 1,835 & 1,188 & Generation \\
      \midrule
      \multicolumn{2}{l}{\textbf{Total}} & \textbf{2,040} & 3--31 & 124 & 2,100 & 588 & \\
      \bottomrule
      \end{tabular}
  \caption{Dataset statistics for ICLEval. Classification (``Clf.'') tasks have a fixed label space; generation tasks require open-ended output. Dictionary (``Dict.'') Search comprises two subtasks (String and Number), which are merged into a single task in the main results (Table~\ref{tab:results}).}
  \label{tab:icleval_stats}
\end{table*}

\section{Experimental Details}
\label{sec:experimental_details}

\subsection{Benchmark}
Table~\ref{tab:icleval_stats} reports dataset statistics for ICLEval~\citep{icleval}. Each task targets a specific facet of in-context learning: \emph{exact copying} tests whether the model can reproduce content from its context via prefix matching, while \emph{rule learning} tests whether it can infer and apply a transformation rule from the demonstrations. Factual entities are replaced with hash strings so that correct predictions require in-context inference rather than recall of pretraining knowledge. Demonstrations are generated dynamically per sample (i.e., each sample has a unique set of exemplars), and all predictions are scored by exact match. For full task descriptions and construction details, we refer the reader to \citet{icleval}.

\subsection{Optimization Pseudocode}
\label{app:algorithm}

Algorithm~\ref{alg:main} presents the pseudocode for the proposed end-to-end in-context calibration.

\begin{algorithm*}[htbp]
  \caption{Self-Improving In-Context Learning}
  \label{alg:main}
  \begin{algorithmic}[1]
  \Require Few-shot prompt $\mathcal{P}$
  \State $X_0 \leftarrow \Call{Embed}{\mathcal{P}}$ \Comment{Initial prompt embeddings}
  \If{$f(X_0) < \tau$}
      \State \Return $X_0$ \Comment{Proxy gate}
  \EndIf
  \For{$k = 0, 1, \ldots, K-1$}
      \State $f_{\mathrm{base}} \leftarrow f(X_k)$ \Comment{Evaluate proxy via forward pass}
      \For{$i = 1, \ldots, N$}
          \State Sample $U_i \sim \mathcal{N}(0, I)$ \Comment{Same shape as $X_k$}
          \State $f_i \leftarrow f(X_k + \mu\, U_i)$ \Comment{Perturbed evaluation}
      \EndFor
      \State $\hat{g}_k \leftarrow \frac{1}{N}\sum_{i=1}^{N} \frac{f_i - f_{\mathrm{base}}}{\mu}\, U_i$ \Comment{Gradient estimate}
      \State $\hat{g}_{k,t} \leftarrow \hat{g}_{k,t} / \max(1,\, \|\hat{g}_{k,t}\|_2)$ for each $t$ \Comment{Clip per token}
      \State $X_{k+1} \leftarrow X_k + \eta\, \hat{g}_k$ \Comment{Ascent step}
      \State $X_{k+1} \leftarrow \Call{CosineProject}{X_{k+1}, X_0, \kappa}$ \Comment{Cosine constraint}
  \EndFor
  \State \Return $X^\star$ \Comment{Embedding that has achieved the highest proxy}
  \end{algorithmic}
\end{algorithm*}

\subsection{Hyperparameters}
\label{app:hyperparams}

We swept $\mu$ and learning rate $\eta$ on a small representative subset of 140 samples. We first tuned on Llama~3.1-8B. After selecting the best configuration, we constructed model-specific grids for the remaining models by scaling these values by $\bar{E}/\sqrt{d}$, where
\[
\bar{E} \;=\; \frac{1}{|\mathcal{V}|}\sum_{i=1}^{|\mathcal{V}|}\|E_i\|_2
\]
is the mean row-norm of the embedding matrix $E \in \mathbb{R}^{|\mathcal{V}| \times d}$ and $|\mathcal{V}|$ is the vocabulary size. We then ran this scaled, model-specific grid to select the best hyperparameters for each model. The final tuned values are listed in Table~\ref{tab:tuned_params}.

\begin{table*}[t]
  \centering
  \begin{tabular}{@{}lccc@{}}
  \toprule
  \textbf{Parameter} & Llama~3.1-8B & Qwen3-4B & Gemma~2-2B  \\
  \midrule
  Perturbation scale $\mu$             & 0.004 & 0.004 & 0.001 \\
  \# Monte Carlo samples $N$           & 16 & 8 & 8 \\
  Stepsize $\eta$                      & 0.05 & 0.06 & 0.035 \\
  Cosine similarity threshold $\kappa$ & 0.2 & 0.2 & 0.2 \\
  Proxy gate threshold $\tau$          & 0.05 & 0.05 & 0.05 \\
  \midrule
  Dimensionality $d$                   & 4096 & 2560 & 2048 \\
  \bottomrule
\end{tabular}
\caption{Model-specific tuned hyperparameters of our method along with the embedding (or hidden) dimensionality. For each model, we use the same set of hyperparameters across all samples.}
\label{tab:tuned_params}
\end{table*}

\subsection{Implementation}
\label{app:implementation}
We implemented the proxy and ran all evaluations using Hugging Face's \texttt{transformers} library~\citep{transformers_hf}. Since no tokens are generated during optimization, the runtime overhead is modest; optimized inference stacks such as vLLM~\citep{vllm} could reduce it further, though as of v0.17, vLLM does not yet support log-probability output for input tokens. Baselines were implemented using the code released by the respective authors.

Our benchmark implementation directly adopts the data and evaluation protocol from the ICLEval codebase\footnote{\url{https://github.com/RUCBM/ICLEval}}. The only model shared with the original ICLEval evaluation is Llama~3.1-8B (compared against the Llama 3-8B results reported in the paper). Most task-level scores agree closely, with ours tending slightly higher. For Count \& Navigation and Format Check, we observe notably lower scores (29\% vs.\ 52\% and 8\% vs.\ 30\%, respectively), which we attribute to behavioral differences between Llama 3 and Llama 3.1---both tasks are documented by the ICLEval authors as highly sensitive to model priors. For String Completion and Dictionary Search, our results (57\% and 89\%) are consistent with the paper's reported values (57\% and 87\%) once one accounts for what appears to be a transposition of those two columns in Table~7 of the original paper, as can be verified by cross-referencing with the grouped scores in Table~2 and the task-to-category mapping in Table~1.

\subsection{Computational Resources}
\label{app:compute}

All experiments were run on a single workstation with two NVIDIA RTX A6000 GPUs (49\,GiB each).

\subsection{Optimization Duration}
\label{app:iterations}

\begin{table}[t]
\centering
\small
    \begin{tabular}{l ccc}
    \toprule
    \textbf{Task} & Llama~3.1-8B & Qwen3-4B & Gemma~2-2B \\
    \midrule
    String Completion    & 51.94 &  52.90 &  32.54 \\
    Dict.\ Search        & 38.59 &  55.42 &   9.17 \\
    \addlinespace[0.7em]
    Format Check         & 59.11 &  81.37 &   0.00 \\
    Format Cloning       & 42.13 &  57.41 &  40.65 \\
    Format Conversion    & 48.80 &  64.84 &  21.60 \\
    \addlinespace[0.7em]
    Order Check          & 43.68 &  49.65 &  33.82 \\
    Order Adjustment     & 47.99 &  47.43 &  29.58 \\
    \addlinespace[0.7em]
    Duplication Check    & 51.07 &  69.97 &  54.13 \\
    De-Duplication       & 52.32 &  55.23 &  43.95 \\
    Count \& Navigation  & 51.61 &  68.26 &  35.50 \\
    Relation Analysis    & 40.29 &  49.74 &  71.01 \\
    \addlinespace[0.7em]
    List Mapping         & 46.42 &  64.66 &  38.57 \\
    \midrule
    \textbf{Average}     & \textbf{47.83} & \textbf{59.74} & \textbf{34.21} \\
    \bottomrule
    \end{tabular}
\caption{Mean number of optimization steps per task and model, averaged over all samples within each task. The corresponding runtimes range from roughly 10 to 60 seconds.}
\label{tab:iteration_counts}
\end{table}

Table~\ref{tab:iteration_counts} reports the mean number of optimization steps per task and model, averaged over all samples within each task. Optimization runs for a maximum of 250 steps with early stopping at a patience of 5. Qwen3-4B averages the most iterations (${\approx}60$), followed by Llama~3.1-8B (${\approx}48$) and Gemma~2-2B (${\approx}34$). This ordering suggests two factors: Llama uses $N{=}16$ perturbations per step (vs.\ $N{=}8$ for Qwen and Gemma), producing better gradient estimates and faster convergence; and Qwen exhibits the broadest downstream improvement (10 of 12 tasks), indicating a richer proxy landscape that sustains optimization longer before early stopping activates. Format Check on Gemma~2-2B is the only cell with exactly zero iterations, confirming that the proxy gate triggers on every sample for this task. The highest per-model iteration counts tend to coincide with the largest accuracy gains---Qwen Format Check (81~steps, $+104.8\%$) and Gemma Relation Analysis (71~steps, $+300\%$)---providing further evidence that the proxy serves as an informative optimization signal where latent capability exists.

\end{document}